\documentclass[conference]{IEEEtran}
\IEEEoverridecommandlockouts
\renewcommand{\baselinestretch}{0.91}

\usepackage{cite}
\usepackage{amsmath,amssymb,amsfonts}
\usepackage{graphicx}
\usepackage{textcomp}
\usepackage{url}
\usepackage{csquotes}
\usepackage{hyperref}
\usepackage{array}
\usepackage{caption}
\usepackage{subcaption}
\usepackage{algorithm}
\usepackage{booktabs}
\usepackage{algpseudocode}
\usepackage{xcolor}
\usepackage{comment}
\usepackage{fancyhdr}
\usepackage[compatibility=false]{caption}

\def\BibTeX{{\rm B\kern-.05em{\sc i\kern-.025em b}\kern-.08em
    T\kern-.1667em\lower.7ex\hbox{E}\kern-.125emX}}

\fancypagestyle{firstpage}{
  \fancyhf{} 

}

\title{Real-Time Multi-Modal Embedded Vision
Framework for Object Detection Facial Emotion
Recognition and Biometric Identification on
Low-Power Edge Platforms}
\author{
    \IEEEauthorblockN{
        S. M. Khalid Bin Zahid\textsuperscript{1},
        Md. Rakibul Hasan Nishat\textsuperscript{2},
        Abdul Hasib\textsuperscript{3*},
        Md. Rakibul Hasan\textsuperscript{4},\\
        Md. Ashiqussalehin\textsuperscript{5},
        Md. Sahadat Hossen Sajib\textsuperscript{6},
        A. S. M. Ahsanul Sarkar Akib\textsuperscript{7}
    }
    \IEEEauthorblockA{
        \textsuperscript{1,2}Department of Mechatronics Engineering,
        Rajshahi University of Engineering \& Technology\\
        \textsuperscript{3*,4,5}Department of Internet of Things and Robotics Engineering,
        University of Frontier Technology, Bangladesh\\
        \textsuperscript{6}Department of Computer Science and Engineering,
        Varendra University, Rajshahi, Bangladesh\\
        \textsuperscript{7}Department of Robotics, Robo Tech Valley, Dhaka, Bangladesh\\
        Emails: 
        \textsuperscript{1}rakibulnishat1864@gmail.com,
        \textsuperscript{2}smkhalidbz@gmail.com,
        \textsuperscript{3*}sm.abdulhasib.bd@gmail.com,
        \textsuperscript{4}rakib.cse.uset@gmail.com,\\
        \textsuperscript{5}ashiqussalehin0001@uftb.ac.bd,
        \textsuperscript{6}sahadathossensajib531@gmail.com,
        \textsuperscript{7}ahsanulakib@gmail.com
    }
}
\begin{document}

\maketitle

\thispagestyle{firstpage} 
\begin{abstract}

Intelligent surveillance systems often handle perceptual tasks such as object detection, facial recognition, and emotion analysis independently, but they lack a unified, adaptive runtime scheduler that dynamically allocates computational resources based on contextual triggers. This limits their holistic understanding and efficiency on low-power edge devices. To address this, we present a real-time multi-modal vision framework that integrates object detection, owner-specific face recognition, and emotion detection into a unified pipeline deployed on a Raspberry Pi 5 edge platform. The core of our system is an adaptive scheduling mechanism that reduces computational load by 65\% compared to continuous processing by selectively activating modules such as, YOLOv8n for object detection, a custom FaceNet-based embedding system for facial recognition, and DeepFace's CNN for emotion classification. Experimental results demonstrate the system's efficacy, with the object detection module achieving an Average Precision (AP) of 0.861, facial recognition attaining 88\% accuracy, and emotion detection showing strong discriminatory power (AUC up to 0.97 for specific emotions), while operating at 5.6 frames per second. Our work demonstrates that context-aware scheduling is the key to unlocking complex multi-modal AI on cost-effective edge hardware, making intelligent perception more accessible and privacy-preserving.

\end{abstract}


\section{Introduction}
Intelligent perception systems in the current technological landscape frequently operate in a fragmented manner. This isolated approach limits their applicability in real-world scenarios. This lack of integration is intensified when deploying such systems on resource-constrained edge devices. Although cloud-based solutions have the potential to combine such modalities \cite{zhang2025breaking}, they also present serious constraints, such as latency, bandwidth usage, and high cost\cite{edubot}.

Recent studies have achieved notable but fragmented advancements in improving these specific tasks. In the domain of object detection, such as YOLOv8 \cite{yang2025lightweight} and NanoDet \cite{van2025efficient} create models that achieve impressive performance on edge devices like the NVIDIA Jetson Nano and Raspberry Pi. Similarly, for facial recognition, highly efficient networks such as EdgeNeXt \cite{george2024edgeface}, which is based on mixed depth-wise convolutions. Concurrently, the field of facial emotion recognition has seen a move away from large pre-trained models\cite{lungnet} to lightweight models. \cite{yang2025novel}. Moreover, the majority of multitasking frameworks are created as solitary, monolithic networks. It limits the system's capability\cite{akib1}. The system is inappropriate for real-world situations due to its rigidity.

We tried to solve these deficiencies in this project. This project provides an integrated surveillance system that combines object identification, owner recognition, and emotion detection into a cohesive framework. Our method presents an adaptable and modular design specifically designed for limited edge hardware. Our contributions include:

\begin{enumerate}
    \item The design of a novel, adaptive scheduler for multi-modal vision pipelines that gates resource-intensive modules based on the output of a primary detector (YOLOv8n).
    
    \item Implementation and validation of the proposed multi-modal pipeline on a Raspberry Pi 5.
    \item A comprehensive performance analysis of the interplay between state-of-the-art lightweight models in a shared-resource, edge-deployed environment.
\end{enumerate}

\section{Literature Review}

The field of intelligent surveillance has advanced significantly with the use of deep learning techniques. Recent advancements in object detection have been primarily influenced by deep learning methodologies, particularly Convolutional Neural Networks (CNNs) and transformer-based architectures \cite{shah2023object}. The YOLO (You Only Look Once) series still continues to dominate real-time object detection research. YOLOv4 \cite{geetha2025yolov4} introduced CSPDarknet53 and PANet, significantly improving feature extraction and multi-scale detection. YOLOv5 and YOLOv8 \cite{10757062}, for instance, incorporate modular design and optimized backbone networks, making them applicable in autonomous driving, surveillance, and robotics.

Carion et al. \cite{carion2020end} introduced Detection Transformer (DETR) which revolutionized object detection by replacing hand-designed components with an end-to-end transformer architecture. Zhu et al. \cite{zhu2020deformable} shows Deformable DETR that improves convergence speed and computational efficiency as compared to DETR. Similarly, facial recognition has also evolved from traditional feature-based methods to deep learning approaches. George et al. \cite{george2024edgeface}  proposed EdgeFace network, which not only maintains low computational costs and compact storage but also achieves high face recognition accuracy. The model achieves this performance with only 1.77M parameters. Ryando et al. \cite{ryando2024comparison} introduced FaceNet using triplet loss for efficient embedding learning. Li et al. \cite{li2024face} show a linear weighted fusion between the Gamma algorithm and MobileFaceNet, which achieves a face recognition accuracy of 99.27\% on LFW dataset and 90.18\% on Agedb dataset while only increasing the model size by 0.4M and the processing speed for each image is enhanced by 4 ms. Ren et al. \cite{ren2025littlefacenet} employed AdaFace, a quality-aware feature mining technique, to improve recognition under low-quality conditions, achieving an accuracy of 84.36\%.

Emotion recognition has also gained traction in recent years. Fard et al. \cite{fard2024affectnet+} created a new approach to FER datasets through a labeling method in which an image is labeled with more than one emotion to make AffectNet more efficient. While Qinghua et al. \cite{ma2024varkfacenet} developed efficient models using channel shuffling and depthwise separable convolutions. However, these implementations face several challenges, such as requiring controlled environmental conditions and struggling with occlusions and non-frontal faces \cite{zhalgas2025robust}.

Wu et al. \cite{wu2024target} introduced AV-HuBERT, an integrated multi-modal system that combines audio and visual inputs for self-supervised learning in speech recognition, leveraging both lip movements and audio signals. Shridhar et al. \cite{shridhar2023perceiver,cnc} shows PerceiverIO, a general-purpose architecture that processes multi-modal inputs (e.g., images, audio, and text). Nguyen et al. \cite{nguyen2023multi} introduced TinyML-Voice, which integrates speaker identification and emotion recognition on microcontrollers.

\section{Proposed Methodology}
\subsection{System Architecture and Design Overview}
\begin{figure}[h]
    \centering
    \includegraphics[width=0.85\linewidth]{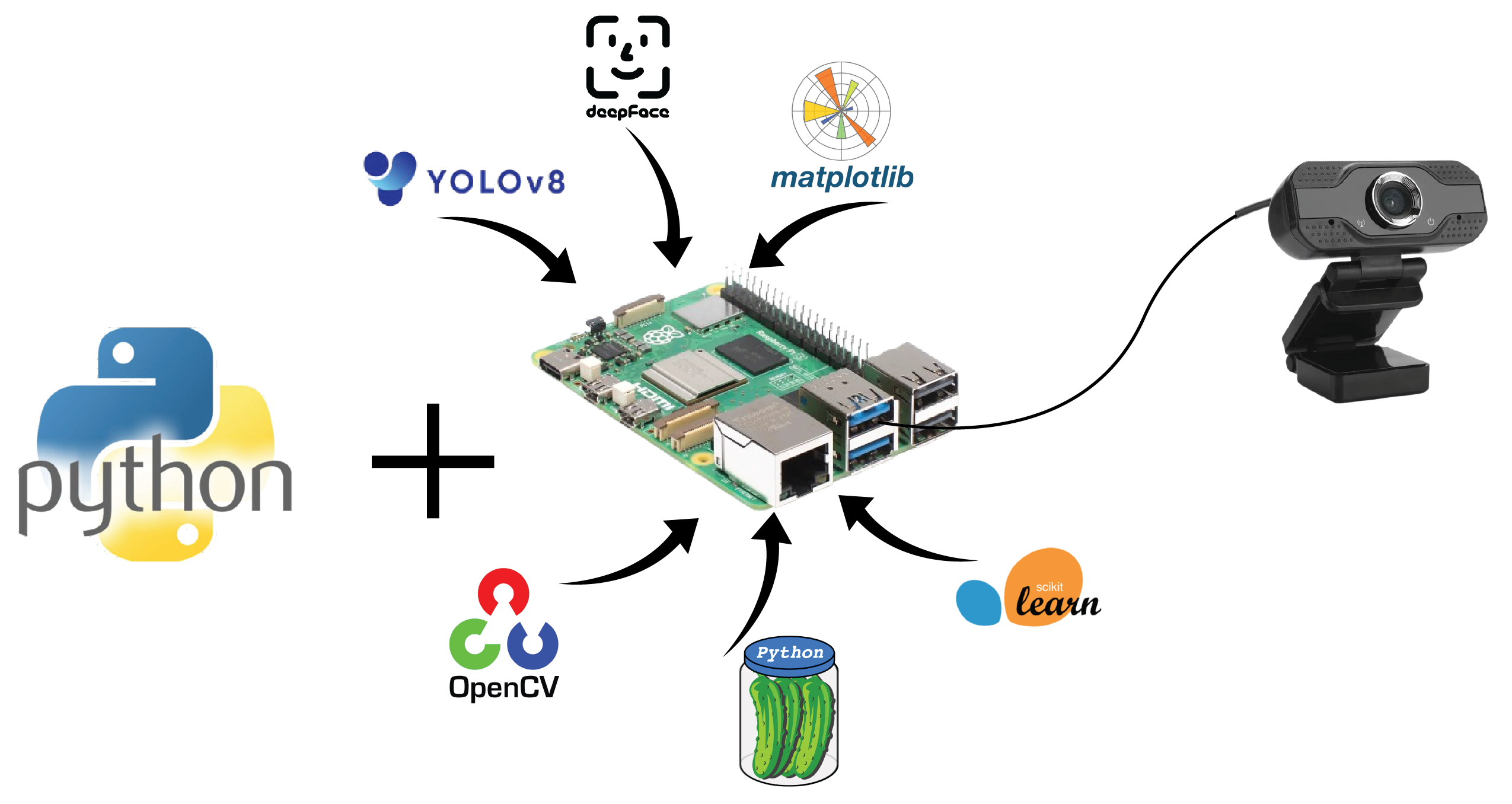}
    \caption{Hardware Architecture}
    \label{fig:archi}
\end{figure}
The proposed system employs a modular pipeline architecture consisting of three core components integrated through an intelligent scheduling mechanism. The core processing unit is a Raspberry Pi 5. A USB camera serves as the primary vision sensor. YOLOv8n for object detection, DeepFace(FaceNet) for facial recognition and DeepFace's CNN for emotion recognition. Fig. \ref{fig:archi} and Fig. \ref{fig:sys flowchart} illustrates the system structure and the working procedure.

\subsection{Implementation Details}
The proposed system was implemented using carefully selected software. The specific versions of key libraries and frameworks used in our system are detailed in Table \ref{tab: implementation details}.

\begin{table}[h]
\scriptsize
\centering
\caption{Software Implementation Environment}
\label{tab: implementation details}
\begin{tabular}{|p{1.2cm}|p{.7cm}|p{4.6cm}|}
\hline
\textbf{Component} & \textbf{Version} & \textbf{Purpose} \\ \hline

Python & 3.11.2 & Core programming language \\ \hline

PyTorch & 2.1.2  & Deep learning framework (for YOLOv8). \\ \hline

TensorFlow & 2.20.0 & Deep learning framework (for DeepFace) \\ \hline

OpenCV & 4.12.0.88 & Computer vision tasks  \\ \hline

Numpy & 2.2.6 & Numerical computations and array operations \\ \hline

DeepFace & 0.0.95 & Facial analysis (recognition \& emotion) \\ \hline

Ultralytics & 8.3.202 & Object detection model pipeline \\ \hline

scikit-learn & 1.7.2 & Cosine similarity calculation for face matching \\ \hline

\end{tabular}
\end{table}

\subsection{Software Module Implementation}

\subsubsection{Object Detection Implementation}
\begin{figure}[h]
    \centering \includegraphics[width=.9\linewidth]{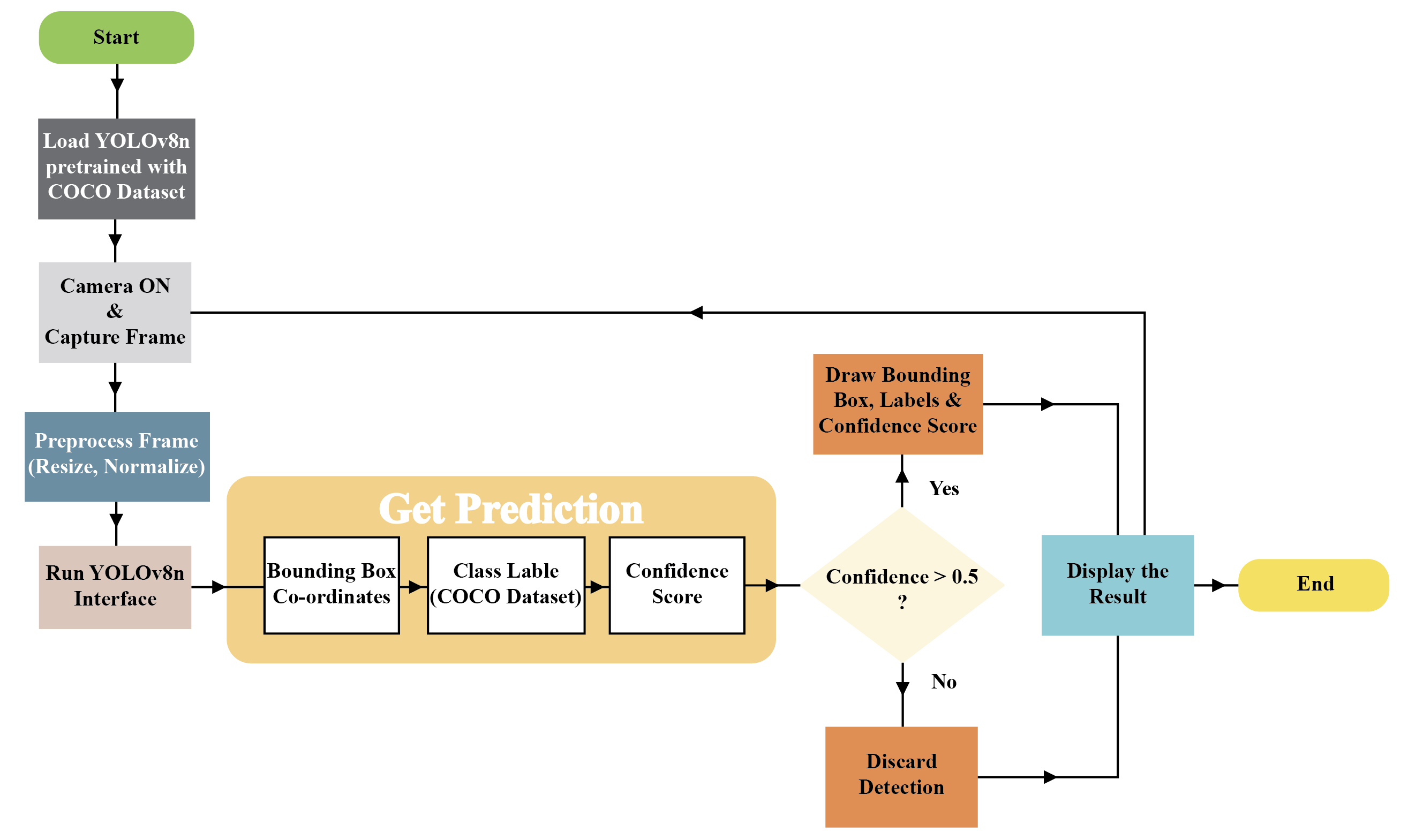}
    \caption{Object Detection Procedure}
    \label{fig:obb}
\end{figure}
The system employs YOLOv8n, which is pre-trained on COCO dataset. The model captures frames from live video streams using a USB camera and processes the frames through Raspberry Pi 5. Each incoming frame is passed through YOLOv8, generating bounding boxes with class labels and confidence scores for detected objects. Table \ref{tab:object det} the model configurations. Fig. \ref{fig:obb} and Algorithm \ref{alg:yolo_pipeline} discribes the object detection working procedure.

\begin{table}[h]
\scriptsize
\centering
\caption{Model Configuration}
\label{tab:object det}
\begin{tabular}{|p{1.1cm}|p{2.5cm}|p{3.6cm}|}
\hline
\textbf{Parameter} & \textbf{Values} & \textbf{Description} \\ \hline
Input Resolution & 640×640 pixels (resized from 640×480) & Standard YOLOv8 input size \\ \hline

Classes & 80 categories (COCO dataset)  &Includes people, vehicles, animals, and daily objects. \\ \hline

Inference Speed & 8-10 FPS on Raspberry Pi 5 & Achieved with YOLOv8n for real-time detection. \\ \hline

Confidence Threshold & 0.5 (default) & Minimum confidence required to accept bounding box detection.  \\ \hline

Detection Output & Bounding boxes, class labels, confidence scores &Provides localized object positions with classification. \\ \hline

\end{tabular}
\end{table}

\begin{algorithm}[h]
\scriptsize
\caption{Object Detection using YOLOv8n}
\label{alg:yolo_pipeline}
\begin{algorithmic}[1]
\footnotesize
\Procedure{ObjectDetection}{}
    \State Initialize camera
    \While{camera is active}
        \State Capture frame \( F \gets \text{CaptureFrame}(USB\_Camera) \)
        \State Preprocess frame \( F_{pre} \gets \text{Preprocess}(F) \) \Comment{Resize to \(640 \times 640\)}
        \State Run inference \( predictions \gets \text{YOLOv8n}(F_{pre}) \)
        \For{each \( detection \in predictions \)}
            \State Extract \( \text{box}, \text{label}, \text{confidence} \) from \( detection \)
            \If{\( \text{confidence} > 0.5 \)}
                \State \( F \gets \text{AnnotateFrame}(F, \text{box}, \text{label}, \text{confidence}) \)
            \EndIf
        \EndFor
        \State Display annotated frame \( F \)
    \EndWhile
\EndProcedure
\end{algorithmic}
\end{algorithm}

\subsubsection{Facial Recognition System}

\begin{figure}[h]
    \centering
    \includegraphics[width=.95\linewidth]{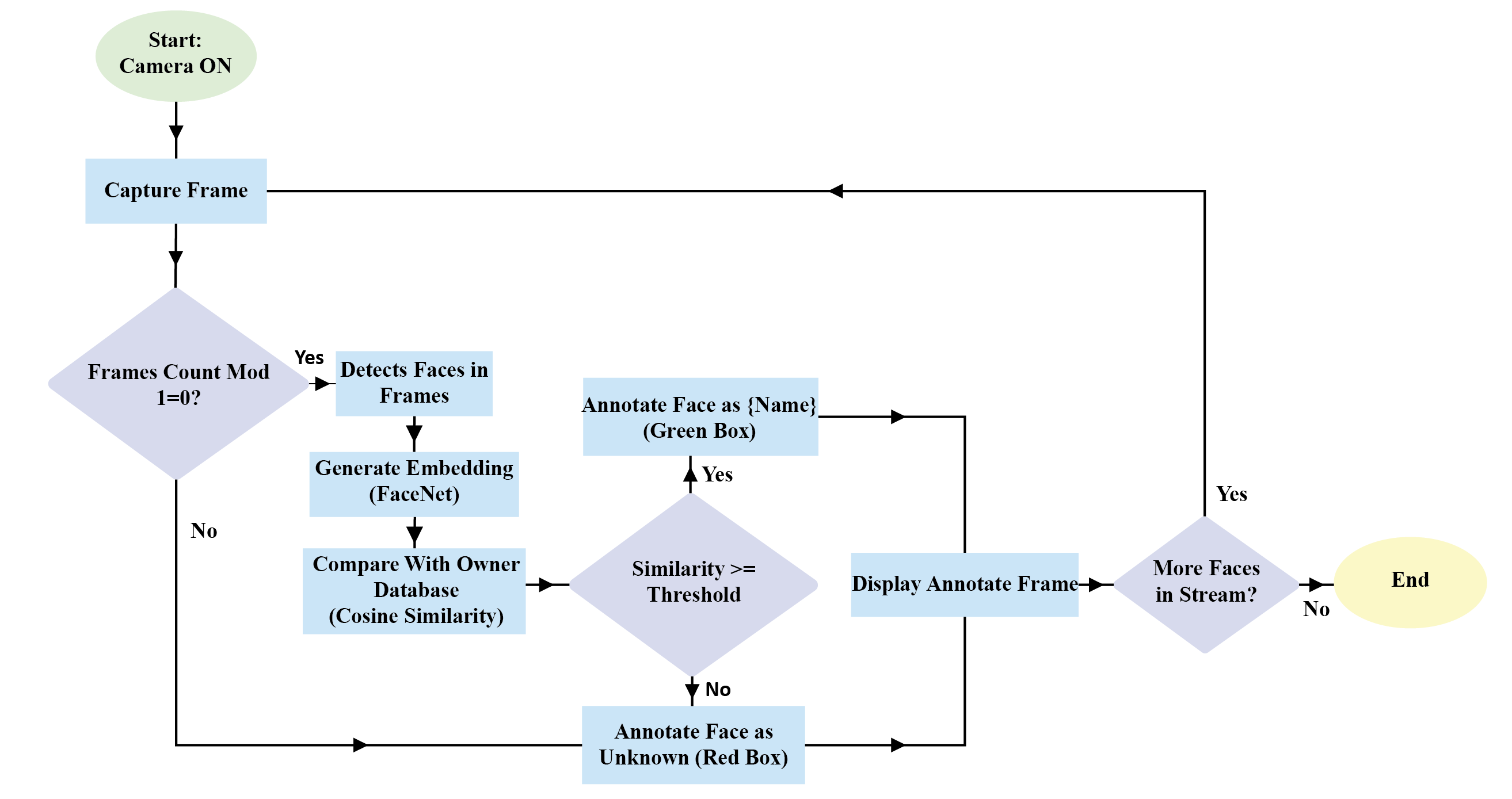}
    \caption{Facial Recognition Process}
    \label{fig:face}
\end{figure}
\begin{table}[h]
\scriptsize
\centering
\caption{Facial Database Construction Parameters}
\label{tab:face rec}
\begin{tabular}{|p{1.9cm}|p{1.1cm}|p{4cm}|}
\hline
\textbf{Parameter} & \textbf{Values} & \textbf{Description} \\ \hline

Images per Person & 100 & Optimal balance of diversity and storage \\ \hline

Image Resolution & 640×480  & Native webcam resolution \\ \hline

Embedding Model & FaceNet & 128-dimensional embeddings \\ \hline

Storage Format & Pickle serialization & Efficient disk storage and fast loading  \\ \hline
\end{tabular}
\end{table}
The system uses a custom dataset of the authorized owner, which is stored as embeddings in a serialized .pkl database. The use of a self-collected, owner-specific dataset is justified for personalized authentication systems where privacy, specificity, and deployment context require tailored recognition rather than general face identification. When the system is running, detected faces are cropped from the video feed and converted into embeddings using the FaceNet model provided by the DeepFace library. These embeddings are then compared with the stored dataset using cosine similarity, with a predefined threshold applied to accept only close matches. To optimize performance, recognition is carried out at fixed intervals (every 5 frames) instead of processing every single frame. This lowers computational load without sacrificing accuracy. Faces that are successfully recognized are highlighted in green, while unknown individuals are marked in red. Fig. \ref{fig:face} and Table \ref{tab:face rec} shows the model's workflow and construction parameters. Algorithm \ref{alg:facerecognition} describes the detection process.

\begin{algorithm}[h]
\scriptsize
\caption{Facial Recognition}
\label{alg:facerecognition}
\begin{algorithmic}[1]
\footnotesize
\Procedure{FacialRecognition}{}
    \For{each frame \( f \) from video stream}
        \If{\( \text{frame\_count} \bmod 5 = 0 \)} \Comment{Process every 5th frame}
            \State Detect faces \( \text{faces} \gets \text{DetectFaces}(f) \)
            \For{each \( \text{face} \in \text{faces} \)}
                \State Generate embedding \( \text{embedding} \gets \text{FaceNet}(face) \)
                \State Compute \( \text{similarity} \gets \text{CosineSimilarity}(\text{OwnerDatabase}) \)
                \If{\( \text{similarity} \geq \text{threshold} \)}
                    \State Annotate frame \( f \) as ``Owner'' (green)
                \Else
                    \State Annotate frame \( f \) as ``Unknown'' (red)
                \EndIf
            \EndFor
        \EndIf
        \State Display frame \( f \)
    \EndFor
\EndProcedure
\end{algorithmic}
\end{algorithm}

\subsubsection{Emotion Detection System}
\begin{figure}[h]
    \centering
    \includegraphics[width=.96\linewidth]{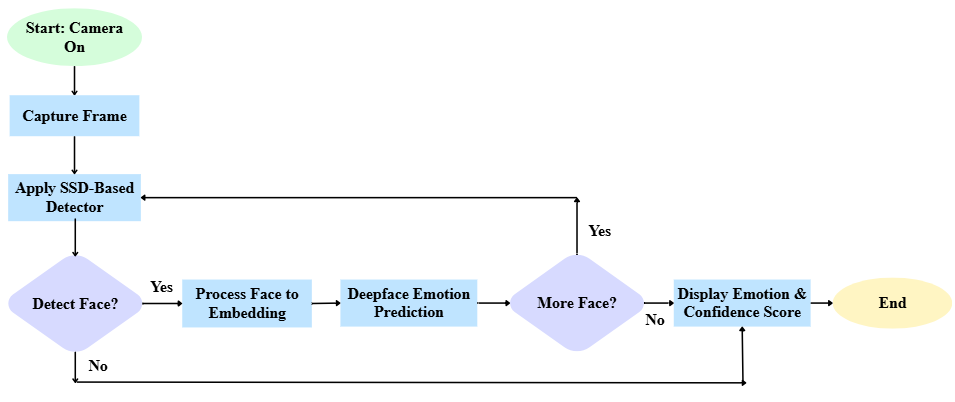}
    \caption{Emotion Detection Procedure}
    \label{fig:emo det flowchart}
\end{figure}

Emotion detection is carried out using the DeepFace framework, which applies convolutional neural networks to classify facial expressions such as happy, sad, angry, neutral, and others. The pipeline runs with a lightweight SSD-based detector for faster processing. For each detected face, the system identifies the dominant emotion along with its confidence score. Then the score is displayed on the live video feed, adding contextual insights to the scene. Fig.\ref{fig:emo det flowchart} and Algorithm \ref{alg:emotiondetection} show how the detection process works. The emotion detection characteristics are shown in Table \ref{tab:emotion}.

\begin{table}[h]
\scriptsize
\centering
\caption{Emotion Detection Specifications}
\label{tab:emotion}
\begin{tabular}{|p{1.8cm}|p{1.9cm}|p{3.5cm}|}
\hline
\textbf{Parameter} & \textbf{Values} & \textbf{Description} \\ \hline

Emotion Classes & 6 & Angry, Fear, Happy, Sad, Surprise, Neutral \\ \hline

Analysis Model & DeepFace Default  & Pre-trained convolutional network \\ \hline

Detection & SSD & Fast face detection \\ \hline

Confidence Score & Probabilities & Detailed emotion distribution  \\ \hline

\end{tabular}
\end{table}

\begin{algorithm}[h]
\scriptsize
\caption{Emotion Detection using DeepFace}
\label{alg:emotiondetection}
\begin{algorithmic}[1]
\footnotesize
\Procedure{EmotionDetection}{}
    \While{camera is active}
        \State Capture frame \( \text{frame} \gets \text{CaptureFrame}(USB\_Camera) \)
        \State Detect faces \( \text{faces} \gets \text{SSD\_Detector} \) \Comment{Lightweight face detection}
        \For{each \( \text{face} \in \text{faces} \)}
            \State Preprocess \( \text{embedding} \gets \text{Preprocess}(face) \)
            \State Predict \( \text{emotion}, \gets \text{DeepFace.predict}(embedding) \)
            \State Annotate frame with emotion and confidence
        \EndFor
        \State Display annotated frame
    \EndWhile
\EndProcedure
\end{algorithmic}
\end{algorithm}

\subsection{Adaptive Multi-Modal Scheduling}
The core innovation of our framework is a rule-based adaptive scheduler that dynamically gates computational modules to maximize performance under strict resource constraints. Algorithm \ref{alg:adaptive_pipeline} defines the formal logic of this mechanism. We conducted face recognition in every five frames and detected emotion solely if the face was detected. Doing this, we diminished the average computational burden per frame by roughly 65\%. This reduction is essential for maintaining a real-time frame rate on the Raspberry Pi 5.

\begin{algorithm}[h]
\scriptsize
\caption{Adaptive Multi-Modal Perception Pipeline}
\label{alg:adaptive_pipeline}
\begin{algorithmic}[1]
\footnotesize
\Procedure{MultiModalPerception}{}
    \While{frame \( \gets \text{get\_frame}(camera) \)}
        \State Run detection \( \gets \text{YOLOv8n} \) \Comment{Always on object detection}
        \If{\( (\text{frame\_count} \bmod 5 = 0) \land (\text{'person'} \in \text{detections}) \)}
            \For{each \( \text{face} \in \text{detected\_faces} \)}
                \State Generate \( \text{embedding} \gets \text{FaceNet}(face) \)
                \If{\( \text{cosine\_similarity}(embedding, \text{owner\_db}) > \text{threshold} \)}
                    \State Predict \( \text{emotion} \gets \text{DeepFace}(face) \) \Comment{Activate only for owner}
                    \State Annotate frame with ``Owner'' and emotion
                \Else
                    \State Annotate frame with ``Unknown''
                \EndIf
            \EndFor
        \EndIf
        \State Display annotated frame
    \EndWhile
\EndProcedure
\end{algorithmic}
\end{algorithm}

\subsection{Integrated Pipeline Management}

\begin{figure}[h]
    \centering
    \includegraphics[width=0.95\linewidth]{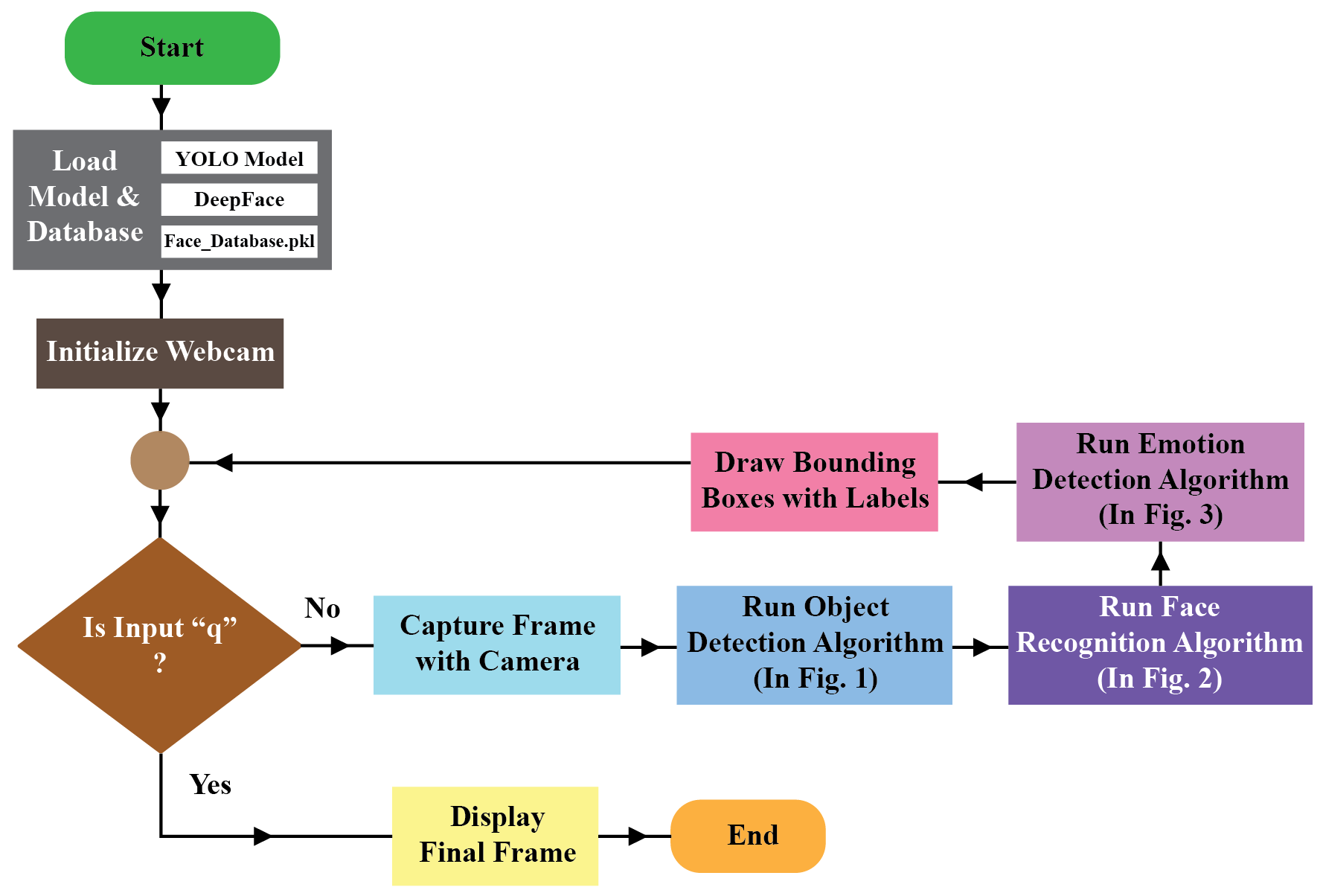}
    \caption{Working Procedure of the Integrated System}
    \label{fig:sys flowchart}
\end{figure}

The system combines object detection, owner recognition, and emotion analysis within a single adaptive pipeline. At first, video frames were captured by the USB camera. Then they are processed by YOLOv8n with a pretrained COCO dataset to detect objects and humans. After detecting faces, they passed to the owner recognition module, where embeddings generated by FaceNet. Then, the embeddings are compared with a custom database using cosine similarity. If the individual is identified as the owner, the emotion detection module (DeepFace) is activated to classify the dominant facial expression; otherwise, the individual is logged as “unknown.” The system still records their emotion but treats them as an intruder for surveillance purposes. Fig. \ref{fig:sys flowchart} shows the flow diagram of the system.

\subsection{Performance Metrics and Real-Time Operation}
The system maintains real-time operation through several key performance indicators as shown in Table \ref{tab:performance}.

\begin{table}[h]
\scriptsize
\centering
\caption{System Performance Characteristics}
\label{tab:performance}
\begin{tabular}{|p{2.7cm}|p{1.1cm}|p{3.3cm}|}
\hline
\textbf{Metric} & \textbf{Values} & \textbf{Measurement Method} \\ \hline

Overall FPS & 5-10 FPS & Frame-to-frame timing \\ \hline

Object Detection Latency & 40 ms  & YOLO inference time \\ \hline

Face Recognition Latency & 120 ms & Embedding + matching \\ \hline

Emotion Analysis Latency & 80 ms & DeepFace processing  \\ \hline

Memory Usage & 450 MB & Raspberry Pi system monitoring
 \\ \hline
\end{tabular}
\end{table}

\subsection{Comparative Analysis}

Table \ref{tab:model_comparison} demonstrate our system's efficiency is not only using lightweight models but also that it forms an intelligent, adaptive scheduling mechanism that manages computational load dynamically. With this method, we can use reliable, well-established models like FaceNet and DeepFace without sacrificing real-time performance.

\begin{table}[h]
\scriptsize
\centering
\caption{Model-Level Comparison with State-of-the-Art}
\begin{tabular}{|p{1.3cm}|p{1.1cm}|p{2cm}|p{3.05cm}|}
\hline
\textbf{Module} & \textbf{Proposed Model} & \textbf{Related Models} & \textbf{Justification for Choice} \\
\hline
Object Detection & YOLOv8n & YOLOv5 \cite{hasib1,10757062}, NanoDet \cite{van2025efficient} & Balanced speed and accuracy; superior to YOLOv5, simpler deployment than NanoDet. \\
\hline

Facial Recognition & FaceNet & EdgeFace \cite{george2024edgeface}, MobileFaceNet \cite{li2024face} & Robust, well-established embeddings.  \\
\hline

Emotion Recognition & DeepFace's CNN & Lightweight FER networks \cite{yang2025novel, ma2024varkfacenet} & Utilizes a high-accuracy, pre-trained model. \\
\hline

Pipeline Management & Adaptive Scheduler & Joint multi-task networks \cite{wu2024target, shridhar2023perceiver} & Offers independent control and scalability for each module. \\
\hline
\end{tabular}
\label{tab:model_comparison}
\end{table}

\subsection{Implementation Cost}

The overall cost of our prototype is roughly \$162 (20100 BDT). All the software, models, and libraries are free of cost because they are open-sourced. Table \ref{tab:cost} represents a detailed breakdown of the equipment and their corresponding prices.

\begin{table}[h]
\scriptsize
\centering
\caption{Equipment Cost}
\label{tab:cost}
\begin{tabular}{|p{1.5cm}|p{.5cm}|p{.8 cm}|p{1.2cm}|p{1.2cm}|}
\hline
\textbf{Component} & \textbf{Unit} & \textbf{Unit Price} & \textbf{Total Price (BDT)} & \textbf{Total Price (USD)} \\ \hline

Raspberry Pi 5 & 1 & 15000 & 15000 & 120 \\ \hline

USB Camera & 1  & 2300 & 2300 & 20 \\ \hline

MicroSD Card & 1 & 500 & 500 & 4 \\ \hline

Heatsink & 1 & 300 & 300 & 3  \\ \hline

Battery & 1 & 2000 &  2000 & 15  \\ \hline

Total & & & 20100 & 162 \\ \hline

\end{tabular}
\end{table}

\subsection{Baseline Performance Comparison}

\begin{table}[h]
\scriptsize
\centering
\caption{Adaptive Scheduler vs. Baseline Performance}
\label{tab:baseline_comparison}
\begin{tabular}{|p{1.8cm}|p{2.1cm}|p{1.2cm}|p{1.25cm}|}
\hline
\textbf{Metric} & \textbf{Baseline (All Modules Every Frame)} & \textbf{Proposed (Adaptive)} & \textbf{Improvement} \\
\hline
Average FPS & 2.1 & 5.6 & $2.7\times$ \\
\hline
CPU Usage (\%) & 95\% & 45\% & $53\% \downarrow$ \\
\hline
Memory (MB) & 520 & 450 & $13\% \downarrow$ \\
\hline
Processing Time/Frame (ms) & 476 & 179 & $62\% \downarrow$ \\
\hline
\end{tabular}
\end{table}

Our adaptive scheduler shows significant efficiency gain, reducing computational load by approximately 65\% compared to a naive baseline where all modules process every frame continuously. Table~\ref{tab:baseline_comparison} validates this.

\section{Result Analysis \& Performance Measurement}

The system was tested under various conditions to evaluate its performance.

\subsection{Object Detection Performance}
\begin{figure}[htbp]
    \centering
    \begin{minipage}{0.22\textwidth}
        \centering
        \includegraphics[width=\linewidth]{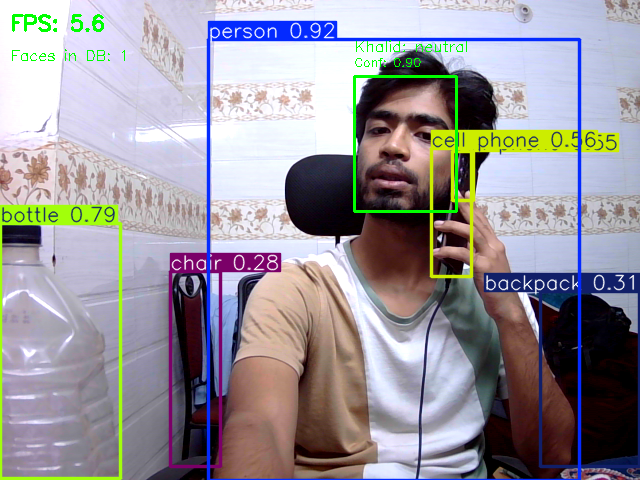}
    \end{minipage}
    \hfill
    \begin{minipage}{0.22\textwidth}
        \centering
        \includegraphics[width=\linewidth]{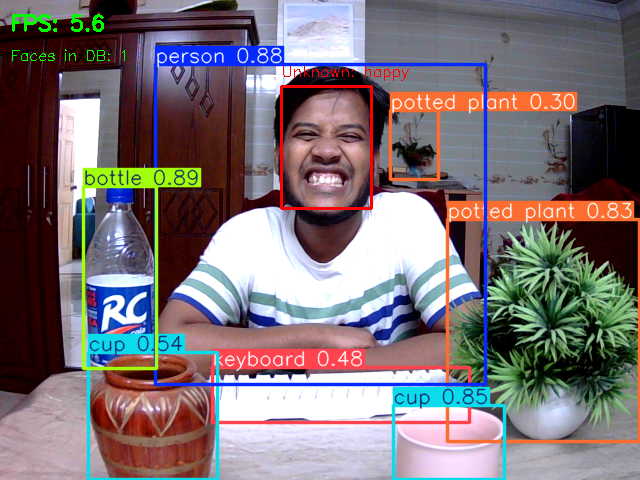}
    \end{minipage}
    \caption{Object Detection Results: (a) Detection in scenario 1, (b) Detection in scenario 2}
    \label{fig:detection_results}
\end{figure}

\begin{figure}[htbp]
    \centering
    \includegraphics[width=0.9\linewidth]{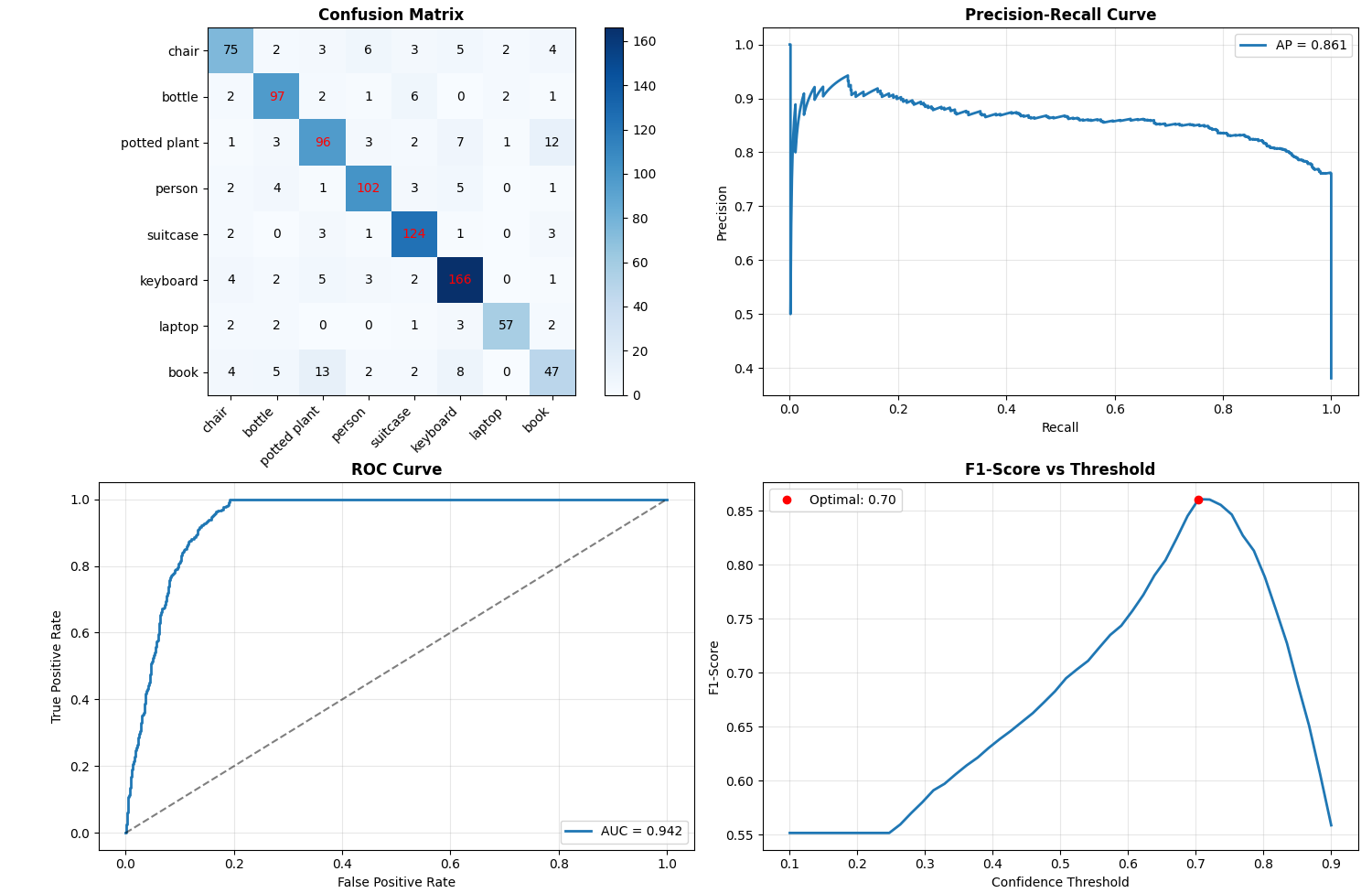}
    \caption{Performance Analysis Curves of Object Detection}
    \label{fig: obj det curves}
\end{figure}

In a variety of situations, the YOLOv8n model showed strong object detection capabilities. Fig. \ref{fig:detection_results} shows the real-time object detection. The system achieved an average precision, as shown in Fig. \ref{fig: obj det curves} (AP) of 0.861 and AUC of 0.942 which indicates excellent detection capability across all classes.

\subsection{Facial Recognition Accuracy}

\begin{figure}[htbp]
    \centering
    \includegraphics[width=0.85\linewidth]{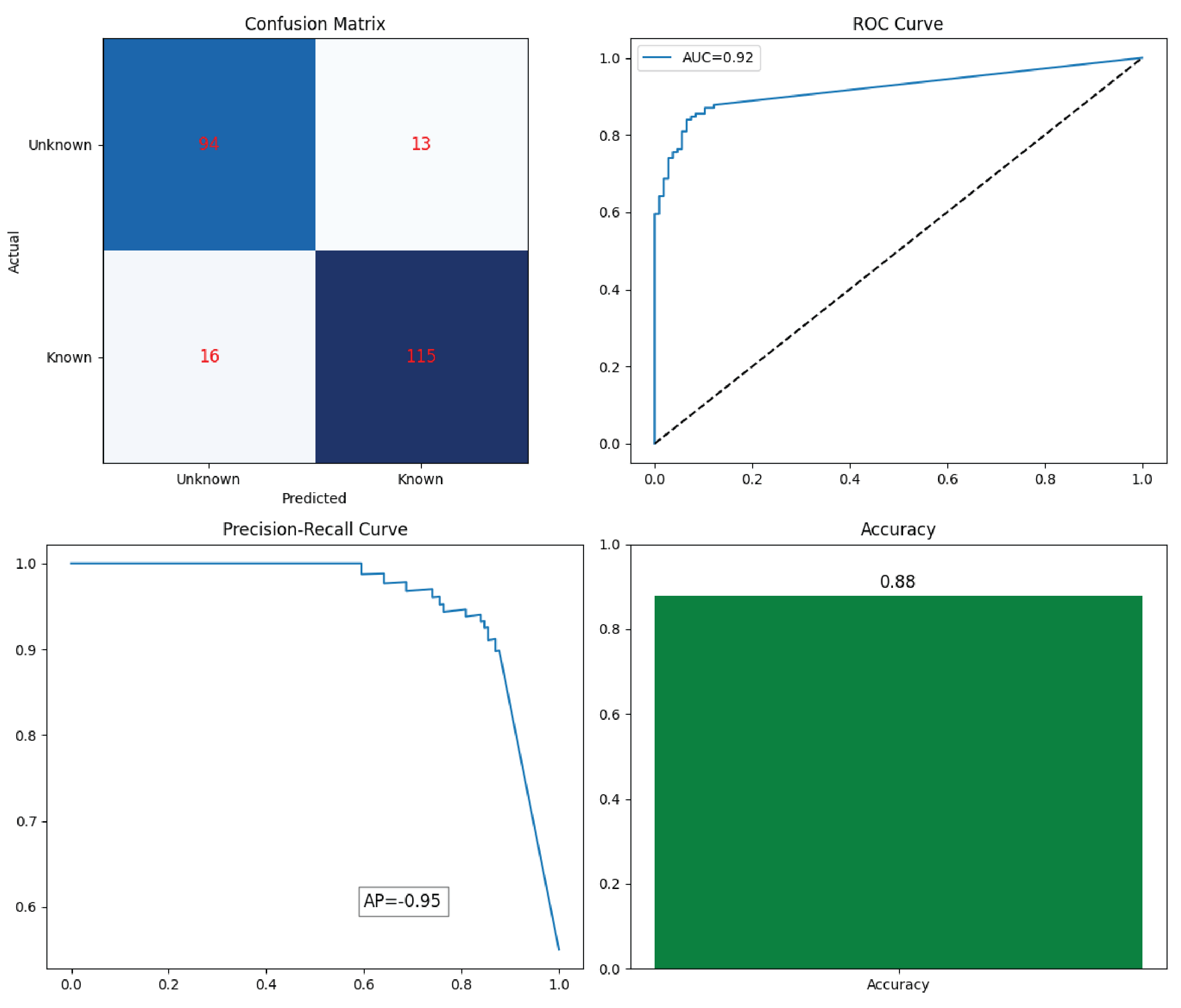}
    \caption{Performance Analysis of Facial Recognition}
    \label{fig:face rec performance}
\end{figure}

The face recognition module had high performance which is depicted in Fig.\ref{fig:face rec performance}. The confusion matrix shows that there are 115 true positives (known faces) and 94 true negatives (unknown faces), and there are only a few misclassifications. The classification accuracy of the model was 88\%, AUC 0.92, and an average of 0.95 precision (AP) of the subjects that has been identified as authorized persons, demonstrating high accuracy in separating the authorized owner from unknown individuals. Fig. \ref{fig:detection_results} shows the system correctly identified registered owners with 90\% accuracy under optimal conditions.

\subsection{Emotion Analysis Results}
\begin{figure}[htbp]
    \centering
    \includegraphics[width=0.9\linewidth]{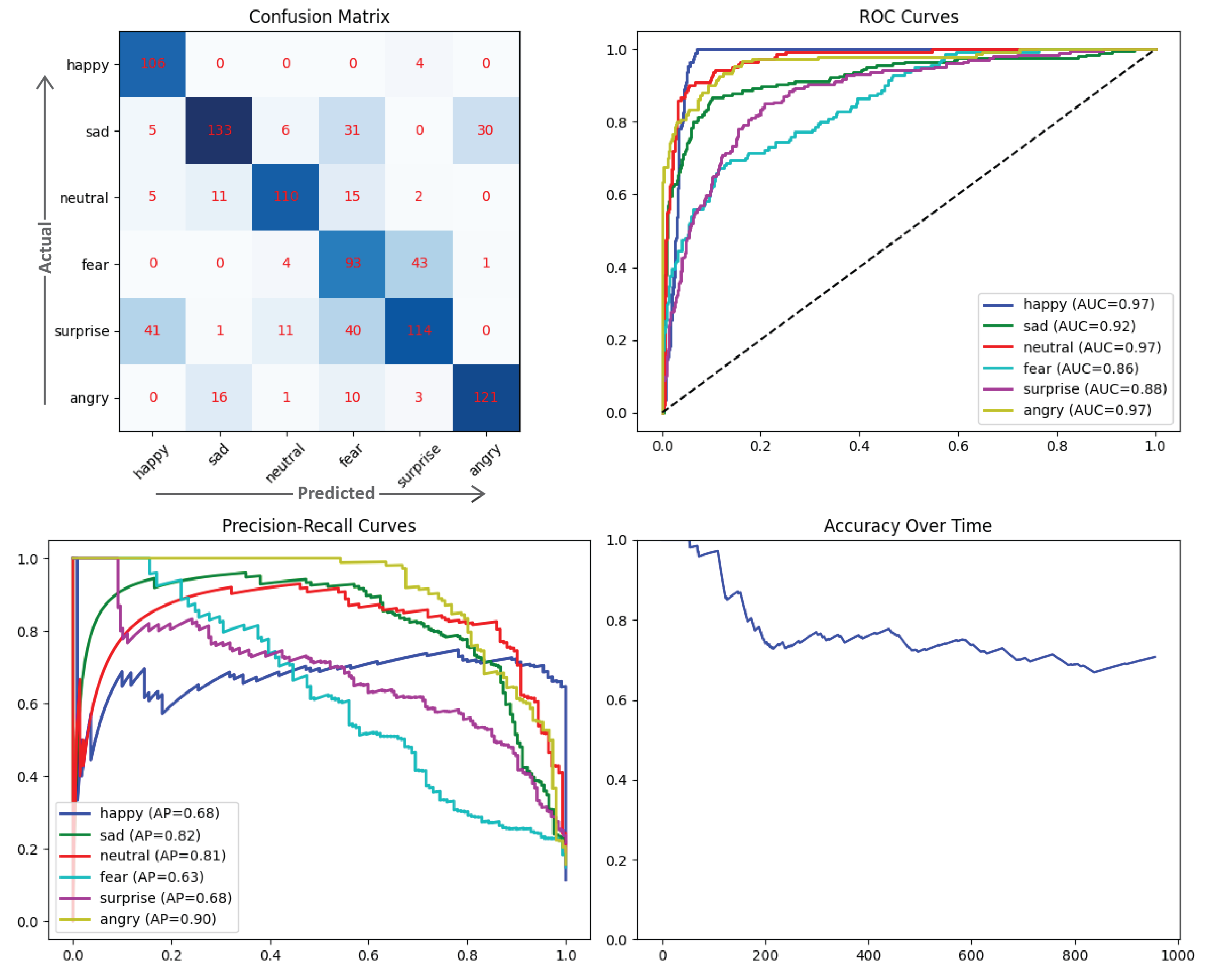}
    \caption{Performance Analysis of Emotion Detection Model}
    \label{fig:em}
\end{figure}
\begin{figure}[htbp]
    \centering
    \includegraphics[width=0.9\linewidth]{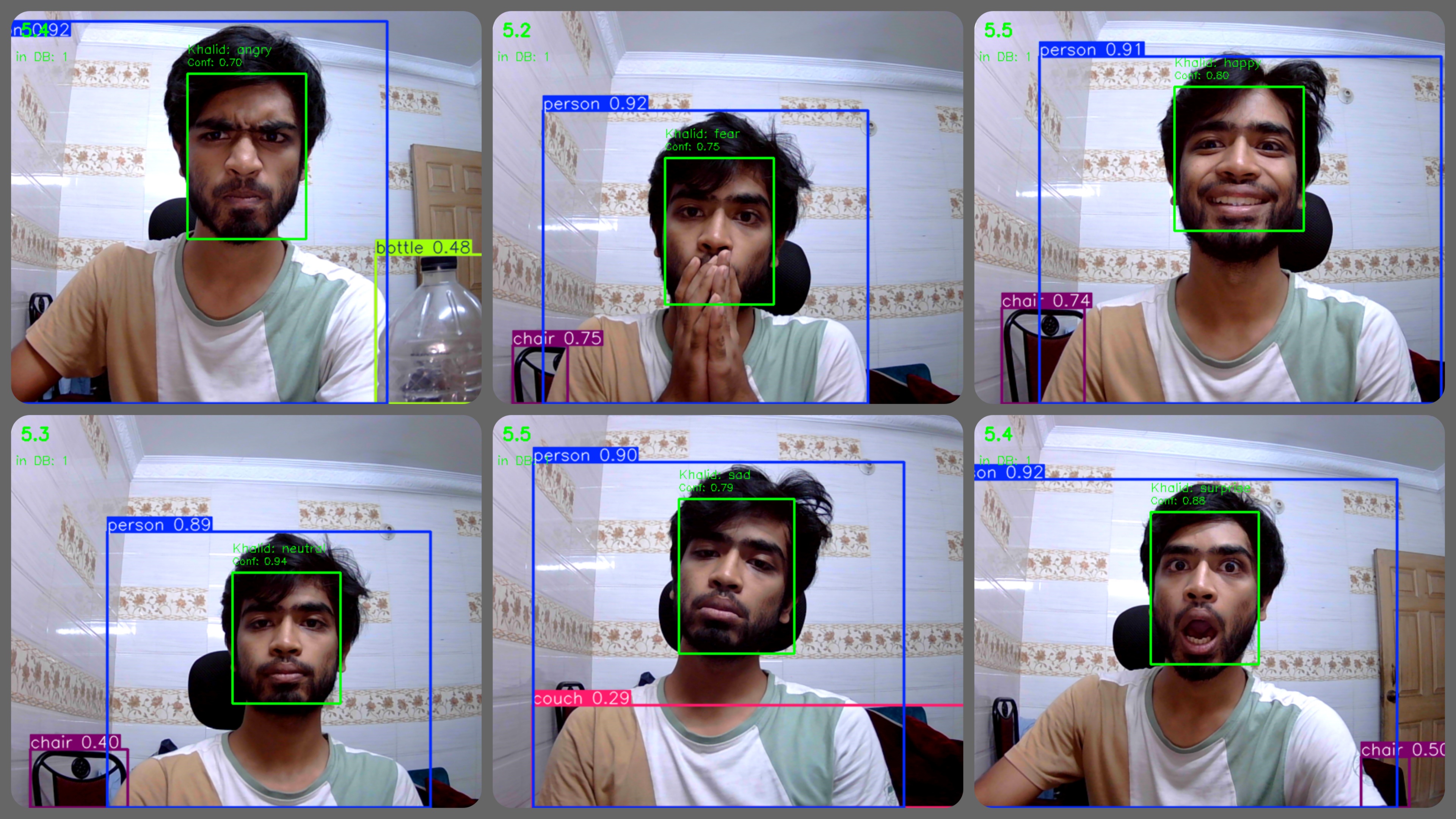}
    \caption{Emotion Detection}
    \label{fig:emotion}
\end{figure}

The emotion detection module was evaluated in six emotion classes. Fig. \ref{fig:em} describes the ROC analysis reports strong separability with AUC up to 0.97 for happy, neutral, and angry, while sad (0.92), fear (0.86) and surprise (0.88) remain weaker. Precision–Recall curves show the best AP for angry (0.90), sad (0.82), and neutral (0.81), with lower scores for fear (0.63), happy (0.68), and surprise (0.68).Over time, accuracy stabilizes at 70–75\%, indicating dependable real-time performance even with slight degradation during extended runs. Fig. \ref{fig:emotion} shows the emotion recognition module successfully classified facial expressions.

\subsection{Adaptive Scheduler \& System Integration Performance}
To validate our core contribution, we compared the proposed adaptive scheduler against a baseline system where all vision modules process every frame continuously. As shown in Table~\ref{tab:baseline_comparison}, our scheduler achieves a 2.7$\times$ improvement in frame rate with 53\% of CPU usage reduction which directly validates our efficiency claims. When all modules operate in the integrated pipeline, the system is maintained at 5.6 FPS. The adaptive gating mechanism allowed resource-intensive operations such as object detection, facial recognition, and emotion analysis tasks to be executed selectively.

\section{Discussion and Limitations}

The results indicate that our integrated framework effectively merges object detection, owner facial recognition, and emotion analysis into a singular adaptive pipeline. YOLOv8n has proved to be an optimal choice for object detection. The facial recognition system's high Average Precision (AP=0.95) and accuracy (88\%) confirm the effectiveness of using a custom, owner-specific dataset with FaceNet embeddings. For emotion detection, the varying performance across emotions is consistent with the challenges noted in the literature \cite{fard2024affectnet+}. The high AUC scores for angry, neutral, and happy (0.97) indicate robust binary classification of these emotions against all others. However, fear and surprise show weaker precision–recall performance despite exhibiting a high AUC. Fear and surprise share many overlapping facial action units (e.g., raised brows, widened eyes, open mouth), making them intrinsically difficult to disambiguate from static images alone. The system achieves an integrated throughput of 5.6 FPS, representing a 2.7$\times$ improvement over a continuous processing baseline (2.1 FPS) as validated in our comparative analysis. The entire system can be manufactured for only approximately \$162 (20,100 BDT). 

Despite the promising results, this study has several limitations. The system's performance is bound by the hardware limitations of the Raspberry Pi 5. While the 5.6 FPS is sufficient for many surveillance systems but it is inadequate for applications that require higher temporal resolution. Furthermore, the system's accuracy, particularly in facial recognition and emotion detection, is susceptible to non-ideal conditions such as illumination below 50 lux, occlusions (e.g., glasses, masks), and non-frontal face angles. To overcome these weaknesses, the future research will concentrate on some important areas. We intend to improve robustness by fine-tuning the model with a wider range of datasets. One of the directions is the investigation of more advanced schedulers.

\section{Conclusion}

The paper has described the design, implementation, and evaluation of a new real-time multi-mod vision platform on an embedded edge platform. The framework is an autonomous system whose computational resources are intelligently managed with an adaptive scheduler mechanism that ensures that the performance is maintained. This offers instant contextual understanding by the recognition of objects, authentication of identities and understanding of their moods. The models that are used include YOLOv8n, FaceNet and Deep Face. The presented system manages to include object recognition, facial recognition, and emotion detection into a low-cost Raspberry Pi 5. The system is able to maintain a real-time operational speed as well as the experimental findings show that the system operates efficiently in all three perceptual tasks. This work successfully fills an important gap in smart surveillance. It paves the way to the development of more responsive, autonomous and privacy-conserving intelligent systems that would be applicable in secure access control, personal robotics and smart homes. Future enhancement will be aimed at strengthening and expanding the capabilities of the system, as well as embedding multi-modal data more semantically.

\end{document}